\relax
\documentclass[letterpaper]{article} 
\usepackage{aaai22}  
\usepackage{times}  
\usepackage{helvet}  
\usepackage{courier}  
\usepackage[hyphens]{url}  
\usepackage{graphicx} 
\usepackage{natbib}  
\usepackage{caption} 
\DeclareCaptionStyle{ruled}{labelfont=normalfont,labelsep=colon,strut=off} 
\usepackage{algorithm}
\usepackage{algorithmic}
\usepackage{hhline}
\usepackage{wrapfig}
\usepackage{subfig}
\usepackage{multirow}
\usepackage{pifont}
\newcommand{\cmark}{\ding{51}}%
\newcommand{\xmark}{\ding{55}}%
\usepackage{newfloat}
\usepackage{listings}
\floatstyle{ruled}
\newfloat{listing}{tb}{lst}{}
\floatname{listing}{Listing}
\nocopyright
\title{Spherical Feature Pyramid Networks For Semantic Segmentation}
\author{
    Varun Anand\equalcontrib,
    Thomas Walker\equalcontrib,
    Pavlos Andreadis\\
    \vspace{5pt}
    \texttt{T.M.Walker@sms.ed.ac.uk} \\
    \textnormal{University of Edinburgh}\\ 
}
\affiliations{
}

\usepackage{bibentry}
\usepackage{svg}

\begin{document}

\maketitle

\begin{abstract}
Semantic segmentation for spherical data is a challenging problem in machine learning since conventional planar approaches require projecting the spherical image to the Euclidean plane. Representing the signal on a fundamentally different topology introduces edges and distortions which impact network performance. Recently, graph-based approaches have bypassed these challenges to attain significant improvements by representing the signal on a spherical mesh. Current approaches to spherical segmentation exclusively use variants of the UNet architecture, meaning more successful planar architectures remain unexplored. Inspired by the success of feature pyramid networks (FPNs) in planar image segmentation, we leverage the pyramidal hierarchy of graph-based spherical CNNs to design spherical FPNs ($S^2$FPN). Our $S^2$FPN models show consistent improvements over spherical UNets, whilst using fewer parameters. On the Stanford 2D-3D-S dataset, our models achieve state-of-the-art performance with an mIOU of 48.75, an improvement of 3.75 IoU points over the previous best spherical CNN.
\end{abstract}

\section{Introduction}

In recent years spherical image data has become increasingly common due to omnidirectional imaging and LiDAR sensors used in autonomous vehicles \cite{autonom, lidar}. However, spherical data is encountered across all disciplines as atomic charge distributions \cite{atomiccharge}, brain activity \cite{brain}, climate patterns \cite{climatedata}, and the cosmic microwave background \cite{HEALPIX, defferrard2020deepsphere}. The task of semantic segmentation frequently appears in these spherical domains, perhaps most notably in the visual systems employed by autonomous vehicles. For planar segmentation, feature pyramids are an established component of state-of-the-art segmentation models \cite{effdet, lin2017feature, MRFN, yolov5}. We introduce a Feature Pyramid Network (FPN) for spherical images. As part of this design, we present an improved scheme to transition spherical signals between pyramid levels, and perform an ablation study over these design choices. 

Early attempts to utilize deep learning models for segmenting spherical images involved mapping them to the Euclidean plane and using well-studied planar, convolutional neural networks (CNN) \cite{su2017learning, kim2016room, fan2017fixation, yu2016large}. Almost exclusively, these methods use the equirectangular projection, mapping the latitude and longitude of the spherical images to Euclidean grid coordinates. However, this mapping introduces polar discontinuities, greatly distorting objects in a scene according to their proximity to the poles \cite{cohen2018spherical}. Objects close to the top and bottom edge of the projection have their scale exaggerated, a distortion which planar segmentation models are known to be sensitive to \cite{MRFN,lin2017feature}. In fact, the introduction of boundaries by planar representations is known to harm classification accuracy of close-by pixels \cite{MRFN}. 

Collectively, these drawbacks have motivated a branch of literature aimed at generalizing successful planar deep learning techniques to operate natively on the sphere. The primary effort has been directed towards designing convolutions that directly consume spherical signals without the need for destructive projections. One of the biggest challenges is the lack of a perfectly symmetrical grid to discretize spherical signals \cite{cohen2018spherical,2005}. Without a spatially consistent notion of a `pixel', there exists no clear way to define a traditional spatial convolution operation. Further still, there is no canonical choice of filter orientation on the sphere. This results in undesirable path-dependent convolutions, since spatial kernels will change their orientation depending on the choice of path taken during convolution \cite{cohen2019gauge}. The seminal work on fully spherical CNNs \cite{cohen2018spherical} led to a line of research on spectral methods. These approaches can bypass the aforementioned challenges by breaking the signal down into spherical harmonics and performing convolution in frequency space as a spectral dot product. However, the repeated computation of Fourier transforms and their inverses are very computationally expensive and scale poorly to high resolution images \cite{cohen2018spherical, esteves2018learning}.

\begin{figure*}
    \centering
    \includegraphics[width = \linewidth]{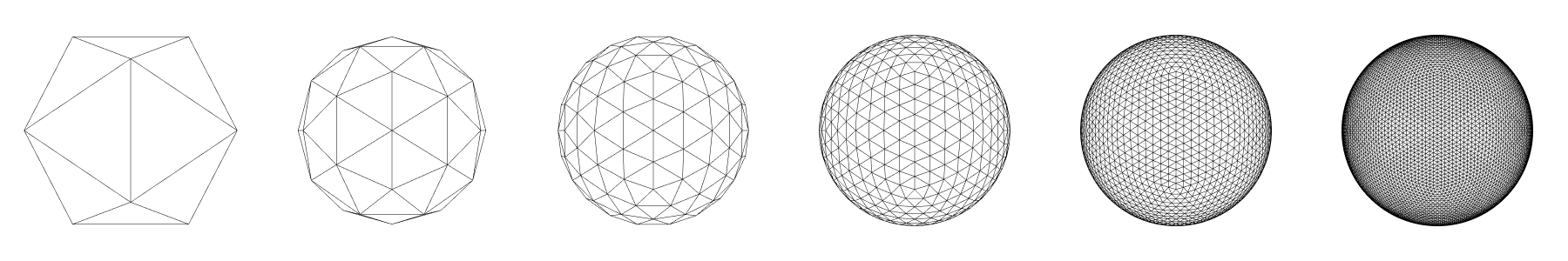}
    \caption{Progressive refinement of icosahedral mesh from $\ell = 0$ to $\ell = 5$.}
    \label{fig:iso}
\end{figure*}

Recently, the works of \cite{jiang2019spherical, defferrard2020deepsphere, shen2022pdoes2cnns} bypass these challenges by sampling the spherical signal onto the nodes of an icosahedral graph, an approximation of a sphere of varying ``levels" (in analogy to image resolution, in the planar case, see figure \ref{fig:iso}). Two of these methods parameterize spherical convolutions as a linear combination of partial differential operators (PDOs). These PDO-based models are efficient and have achieved competitive results on benchmark segmentation data sets \cite{jiang2019spherical, shen2022pdoes2cnns}. 

We continue the trend of generalizing successful planar techniques to the spherical domain. In planar segmentation models, the importance of multi-scale features is long established, even before the success of deep CNNs \cite{hog, surf}. \citet{lin2017feature} introduced Feature Pyramid Networks (FPNs), that have now become common-place in state-of-the-art networks \cite{yolov5, effdet, MRFN, lin2017feature}. FPNs leverage the intrinsic receptive field hierarchy of a CNN to extract semantic feature maps across various scales, which enable them to detect objects of varying shapes and sizes. For our approach, we leverage the varying receptive field of PDOs on an icosahedral graph to design Spherical Feature Pyramid Networks for graph-based spherical models. 

Our contributions are as follows:
\begin{enumerate}
    \item We design spherical feature pyramid networks and present results for a range of model complexities. Our best model improves the state-of-the-art on Stanford 2D-3D-S \cite{2d3ds} by a significant margin.
    \item We present optimal pooling and up-sampling routines for constructing icosahedral feature pyramids. 
\end{enumerate}

\section{Related Work} 

\subsection{Graph-based Spherical CNNs}
Despite the significant progress in making spectral approaches more scalable \cite{kondor2018clebschgordan,cobb2021efficient, mcewen2022scattering}, graph-based spherical CNNs still emerge as the most efficient approach to processing spherical signals. However, a notable cost of approximating the sphere as graph is the diminished ability to facilitate rotational equivariance. The most recent research efforts have been in attaining more uniform spherical samplings, and designing rotationally equivariant convolutions \cite{Perraudin_2019, defferrard2020deepsphere, shen2022pdoes2cnns}. Hence, these works focus on the signal domain and the layer-wise operations, not considering improvements to the architectural design they employ for the task of segmentation. Our work is orthogonal to surrounding research, and builds on the PDO-based approach introduced by \cite{jiang2019spherical} to utilize feature pyramids for segmentation.
\subsection{Multi-scale Features}
Since objects in the real world have different shapes and sizes, it is desirable for networks to be able to extract multi-scale features that capture information across all object scales. Even before the success of deep CNNs, the literature on planar image segmentation and object detection has a long history of methods using Gaussian image pyramids \cite{gaussian_pyramids} to identify objects across scales. These image pyramids were combined with hand-engineered features \cite{hog, scale_invariant_keypoints} to achieve state-of-the-art results at the time. Though CNNs have replaced engineered features, the need for multi-scale features still remains. Early CNN-based architectures \cite{overfeat, ssd} used the pyramidal feature hierarchy of a CNN to do object detection. However, these networks weight semantic information unequally across the levels of the pyramid, and perform poorly on small objects. Feature Pyramid Networks (FPN) \cite{lin2017feature} was the seminal work to introduce a top-down pathway, to extract high-level semantic information uniformly at multiple scales. FPNs simultaneously leverage a CNNs receptive field and semantic hierarchies efficiently, and have been widely adopted in succesful planar object detection and segmentation systems \cite{kirillov2019panoptic}, \cite{effdet}.

\section{Preliminaries}
In this section we provide an overview of the core mathematical components of graph-based spherical CNNs which use PDOs.  

\subsection{Icosahedral Mesh}
We define our signal domain, the icosahedral spherical mesh. As originally proposed by \cite{doi:10.1137/0722066}, we can accurately discretize a sphere by recursively applying a subdivision routine initialized on an icosahedron. In each iteration we add vertices at the midpoints between each pre-existing vertex, and then re-project then to be unit distance to the origin. Fully connecting the new vertices subdivides each original face into four new triangles. Following this scheme we can naturally define upsampling and downsampling algorithms which are analogous to different resolutions in planar images, see figure \ref{fig:iso}. We refer to the original icosahedron as a level-0 ($\ell = 0$) mesh. Each iteration of this subdivision routine increases the level $\ell += 1$. The number of vertices $n_v$ scales with level $\ell$ according to,
\begin{equation}
    n_v = 10\cdot4^\ell + 2.  
\end{equation}
Our input and output spherical signals are level-5 meshes, with the minimum level-0 mesh at the lowest level of the network. 
\subsection{MeshConv Operation}
We apply the "MeshConv" operation as defined in \cite{jiang2019spherical}. Given the spherical signal $F$, and a partial differential operator (PDO) kernel $G_{\theta}$, we parameterize a convolution at each vertex as,
\begin{equation}
    F \ast G_\theta = \theta_0 I F + \theta_1 \nabla_x F + \theta_2 \nabla_y F + \theta_3 \nabla^2 F,
\end{equation}
where $I$ is the identity operator and $\nabla_x$ and $\nabla_y$ are gradients in the east-west and north-south directions respectively. Collectively, the constituent PDOs capture diffusion properties of the signal at each vertex. First and second differential operators, as well as the Laplacian operator, and computed using the Libigl library \cite{libigl}, and follow from results in discrete differential geometry \cite{crane}. Namely, we represent scalar functions on a mesh as a piece-wise linear function with values defined at each mesh vertex:
\begin{equation}
    f(\mathbf{x}) \approx \sum\limits_{i=1}^n \phi_i(\mathbf{x})\, f_i,
\end{equation}
where $\phi_i$ is a piece-wise linear basis function defined on the mesh. For each triangle $\phi_i$ is a linear function which is one only at vertex $\textbf{x} = v_i$ and zero otherwise.

For future discussion, note that with this prescription we are making the choice to perform bilinear interpolation to find the function values between the vertices of the spherical signal. We can consider gradients of piecewise linear functions as simply sums of gradients of the hat functions:
 \begin{equation}
     \nabla f(\mathbf{x}) \approx
 \nabla \sum\limits_{i=1}^n \phi_i(\mathbf{x})\, f_i =
 \sum\limits_{i=1}^n \nabla \phi_i(\mathbf{x})\, f_i.
 \end{equation}
Around a given vertex $v_i$, these basis gradients $\nabla\phi_i$ are zero everywhere on the mesh except the surrounding faces which contain $v_i$. For the spherical gradient signal itself, we sum the gradients on these faces weighted by each face's area. Following \cite{jiang2019spherical}, at each vertex we take the dot product of this gradient with north-south and east-west basis vector fields, separating these components for individual parameterization within the PDO kernel.

Within this framework, the signal's Laplacian can similarly be computed using the cotangent formulation of Laplace-Beltrami operator,
\begin{equation}
    \Delta f(v_i) =  \frac{1} {2A_i} \sum\limits_{{v_j \in \mathcal N(v_i)}} { (\cot \alpha_{ij} + \cot \beta_{ij}) (f(v_j) - f(v_i))},
\end{equation}
where $\mathcal{N}(v_i)$ is the set of vertices in the 1-ring neighborhood of $v_i$, $A_i$ is cell area of vertex $v_j$ and $\alpha_{ij}, \beta_{ij}$ are referred to as the cotangent angles.

\begin{figure}[ht]
    \centering
    \includegraphics[width=4.5cm]{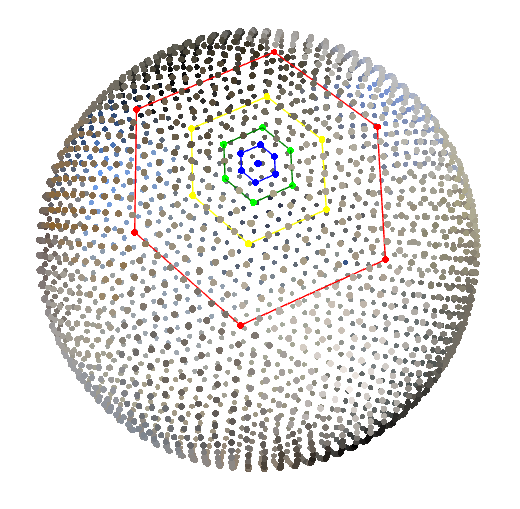}
    \caption{The scale of the the one-ring neighbourhood on level-1 (red), level-2 (yellow), level-3 (green) and level-4 (blue) meshes, displayed on a level-4 icosahedral signal.}
    \label{fig:receptivefield}
\end{figure}

Note that these computations use information from the one-ring neighbourhood of each vertex to compute features. The size of this neighbourhood is larger at lower levels of the mesh (see Figure \ref{fig:receptivefield}). This property enables MeshConv operating at lower mesh levels to have a larger receptive field on the input spherical signal. We leverage this receptive field hierarchy to construct Spherical Feature Pyramid Networks ($S^2$FPNs).
\section{Methodology}

In this section, we describe the architecture of our $S^2$FPN model, followed by a description of our design choices and their impact on the receptive field of the models. 

\subsection{Feature Pyramid Networks}
\begin{figure}
    \centering
    \includegraphics[width=0.85\linewidth]{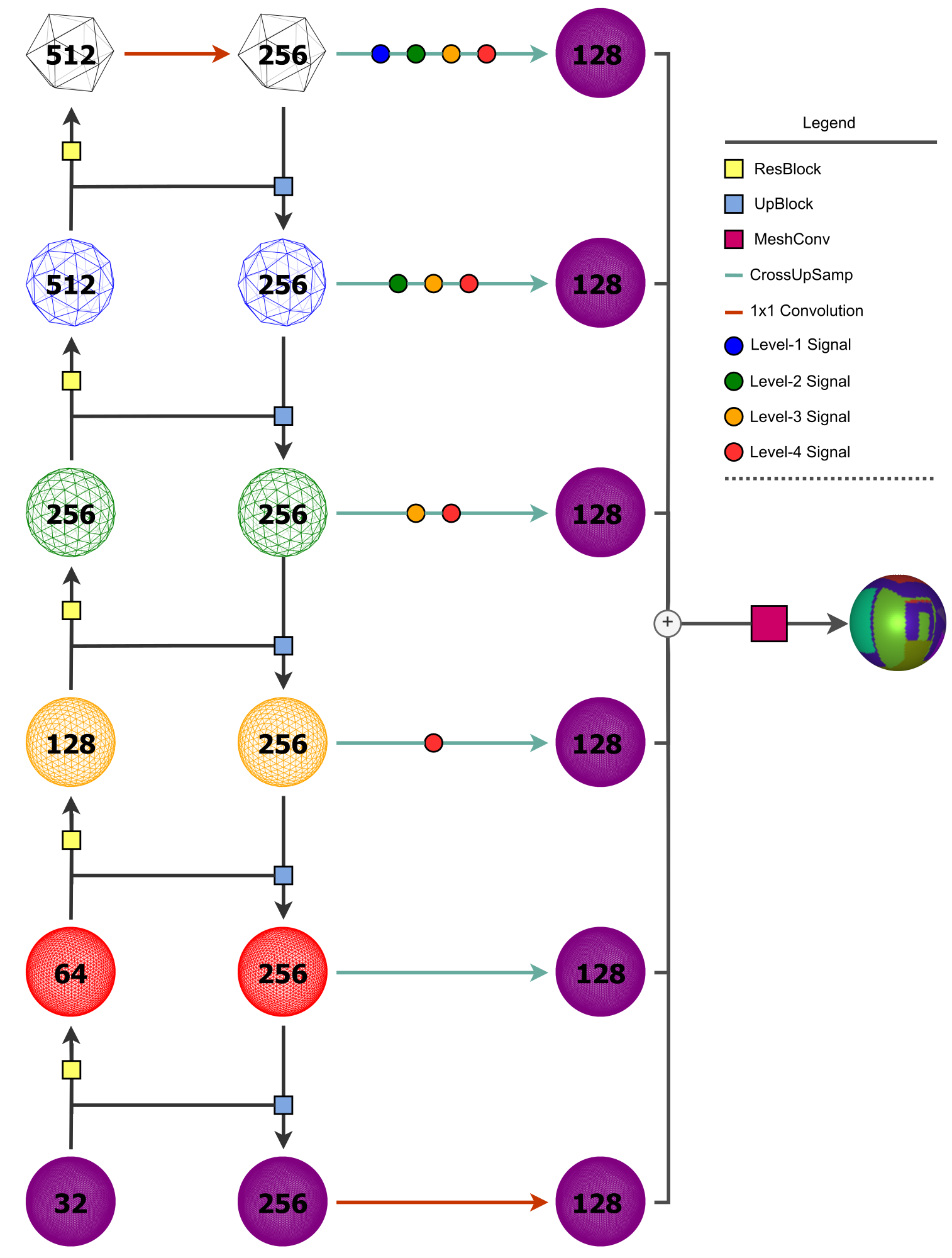}
    \caption{Overview of the L0:5 model used for the Stanford 2D-3D-S dataset. The numbers represent the number of channels at that level.}
    \label{fig:spherical_arch}
\end{figure}

Our spherical FPN consists of three stages which we detail below. Figure \ref{fig:spherical_arch} provides an overview of our overall architecture.

\textbf{Encoder}: We use a modified version of the encoder used by the Spherical UNet \cite{jiang2019spherical}. Starting with a level-5 mesh, we apply bottlenecked ResBlocks similar to until the lowest mesh level of the model. Each ResBlock consists of a MeshConv operation sandwiched between standard 1x1 convolutions, with average pooling applied after the MeshConv. Our model differs from the UNet in terms of the order of operations and the choice of downsampling. This is explained further in section \ref{UpDown}. The number of channels is set to 32 at level-5 and is doubled on each transition to a lower level mesh. At level-0, the channel width is capped at 512 (equal to level-1), to avoid a drastic increase in parameters. Such a design results in a semantic hierarchy, with lower mesh levels extracting higher-level semantic information, at coarser resolutions. 

\textbf{Pyramid}: To construct our feature pyramid we follow \cite{lin2017feature} and add a 256-channel top-down pathway with lateral connections from the encoder. This pathway upsamples features from the lowest level of the encoder, building semantically rich feature maps across all scales. Each upsampling block consists of bilinear upsampling, followed by addition with feature maps of the same level from the encoder (Upsamp Block). The lateral connections use a 1x1 convolution to match the number of channels in the pyramid. By keeping the number of channels fixed across all pyramid levels, we treat each scale equivalently, weighting semantic information equally across all scales. 

\textbf{Head}: While the design of the encoder and the pyramid are generic and can be used for any task, we use a classification head designed for semantic segmentation following \citet{kirillov2019panoptic}. Starting from the lowest mesh level, every pyramid feature map is upsampled using bilinear upsampling followed by MeshConv, to the output mesh level (CrossUpSamp). This results in a set of level-5 feature maps which are element-wise summed and passed through a final MeshConv layer for prediction. We set the the number of channels to be 128 across all the head feature maps.

Finally, all our MeshConv operations are followed by batch normalization and ReLU. Our pyramid differs from \citet{lin2017feature}, who empirically found non-linearities to not have an effect on their task. We found batch normalization to be necessary to prevent explosions in MeshConv’s differential operators.

\subsection{Up/Down-sampling Spherical Signals}
\label{UpDown}
Between levels of the feature pyramid the spherical image is up/down-sampled. We diverge from the seminal work on PDO-based spherical CNNs \cite{jiang2019spherical} in how we compute new vertex values when transitioning between levels, as well as the order of operations in our ``Up/Down-Samp" blocks. 
\subsubsection{Down-sampling}
\cite{jiang2019spherical} downsample by sampling the signal values only at the nodes shared by the subsequent, lower-level graph. Instead, we average pool vertex values in a 1-ring neighborhood around these shared vertices to ensure no information is lost. By taking information from a 1-ring neighbourhood, we also increase the receptive field of vertices at each level.
\subsubsection{Up-sampling}
\begin{figure}[ht]
    \centering
    \subfloat[\centering Padded upsampling. ]{{\includegraphics[width=0.3\linewidth]{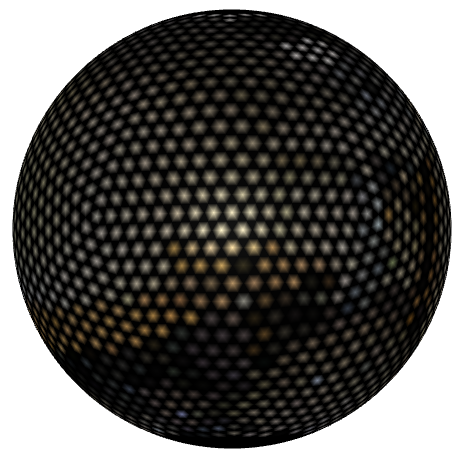}}}%
    \quad
    \subfloat[\centering Bilinear upsampling. ]{{\includegraphics[width=0.3\linewidth]{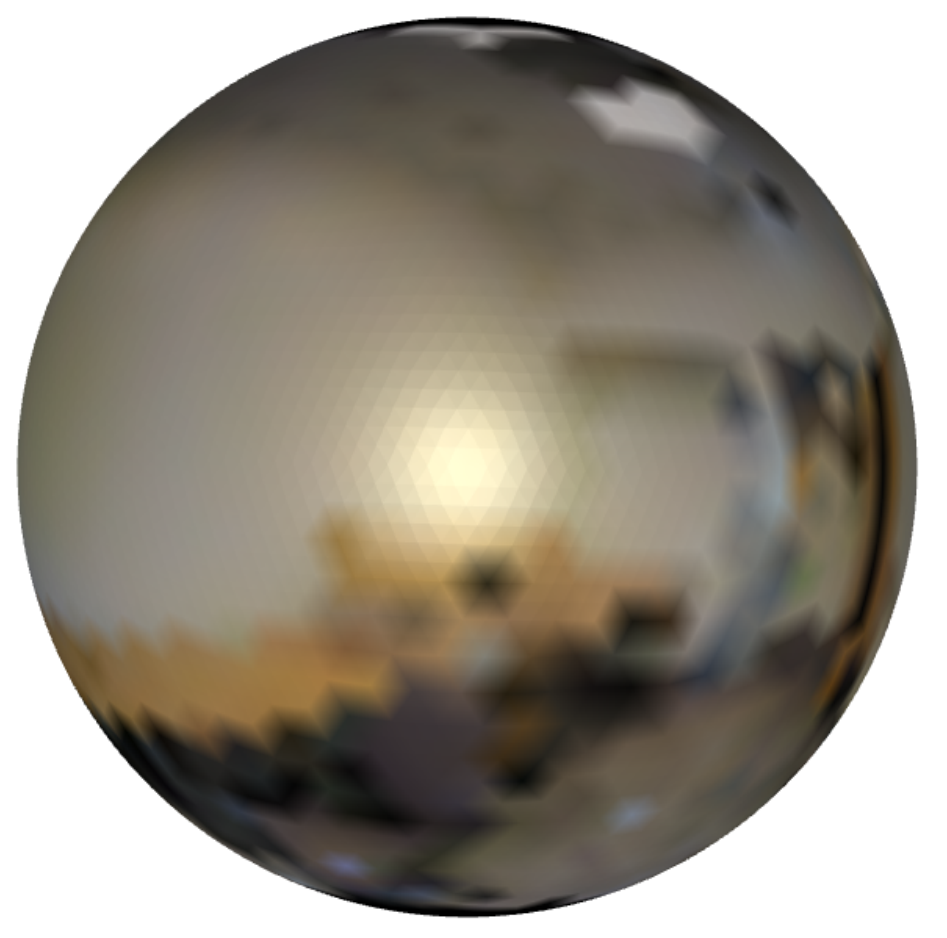}}}\\%
    \quad
    \subfloat[\centering MeshConv applied to padded upsampling.  ]{{\includegraphics[width=0.3\linewidth]{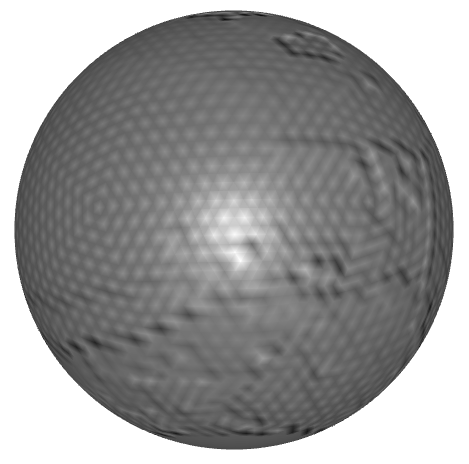}}}
    \quad
    \subfloat[\centering MeshConv applied to bilinear upsampling. ]{{\includegraphics[width=0.3\linewidth]{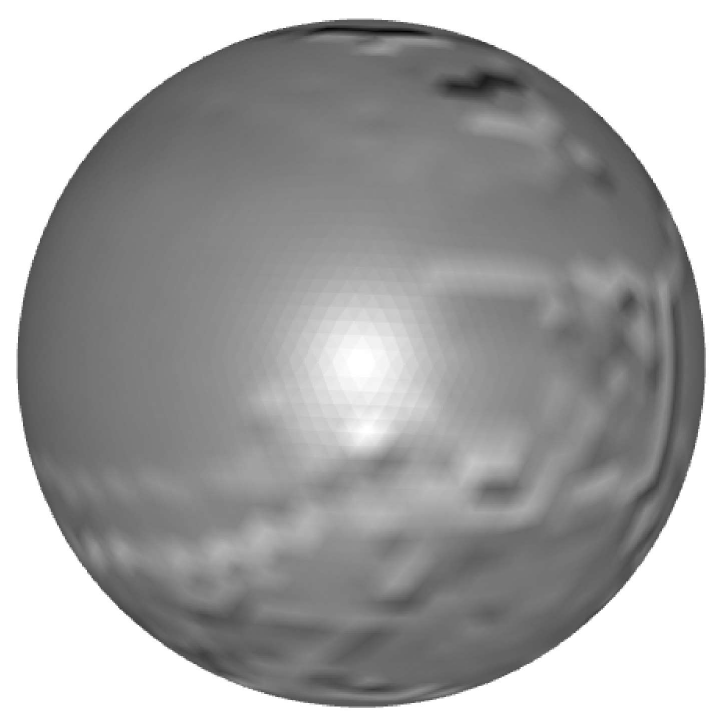}}}
    \caption{Figure a) and b) show the results of upsampling a level-3 signal to level-4 with zero-padded and bilinear upsampling respectively. Figures c) and d) show the output of a single MeshConv operation, with all parameters set to 1. We can see unwanted ``dotted" artifacts present in the zero-padded example.}
    \label{fig:upsampling}
\end{figure}

In each iteration of the icosahedral sub-division routine, additional vertices are added at the midpoints between existing vertices. In the seminal model proposed by \cite{jiang2019spherical}, new vertices are assigned a value of zero, essentially ``zero-padding" the shared vertices of the existing, lower-level, signal. This approach introduces artificial edges, since every vertex in the mesh is now surrounded by a ring of black vertices (Figure \ref{fig:upsampling}a). The MeshConv operator is particularly sensitive to these artifacts, since the constituent differentials act as edge detectors. Accordingly, we bilinearly interpolate the signal at parent vertices in order to compute the new midpoint vertex values.  
\subsubsection{Order Of Operations} 
Finally, note that as part of the up/down-sampling block, \cite{jiang2019spherical} apply a final MeshConv after performing up/down-sampling. In our approach, this order of operations is swapped such that convolution is performed prior to transitioning between graph levels. \\

\begin{figure}[]
    \centering
    \subfloat{{\includegraphics[width=0.4\linewidth]{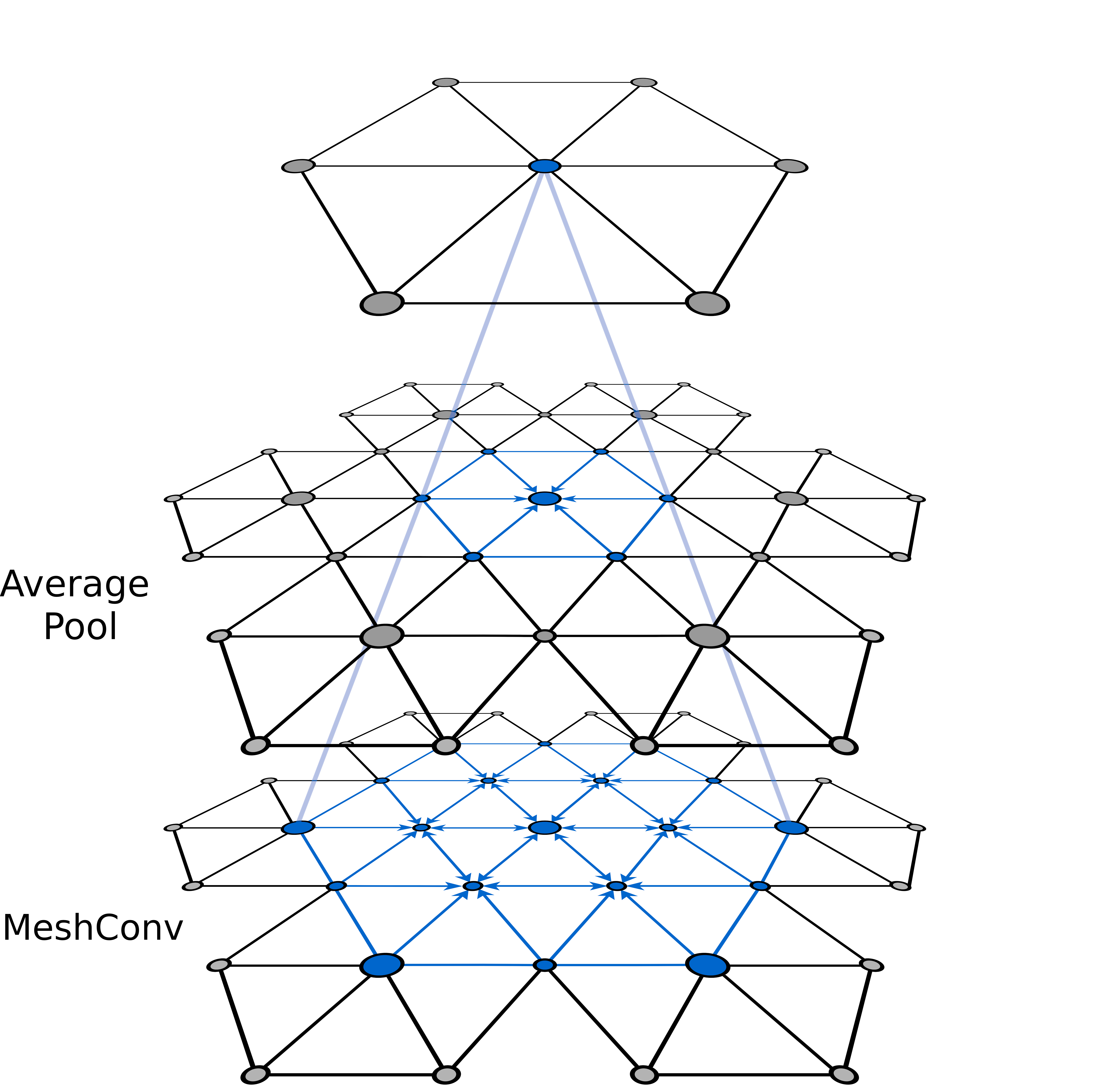}}}%
    \quad
    \subfloat{{\includegraphics[width=0.44\linewidth]{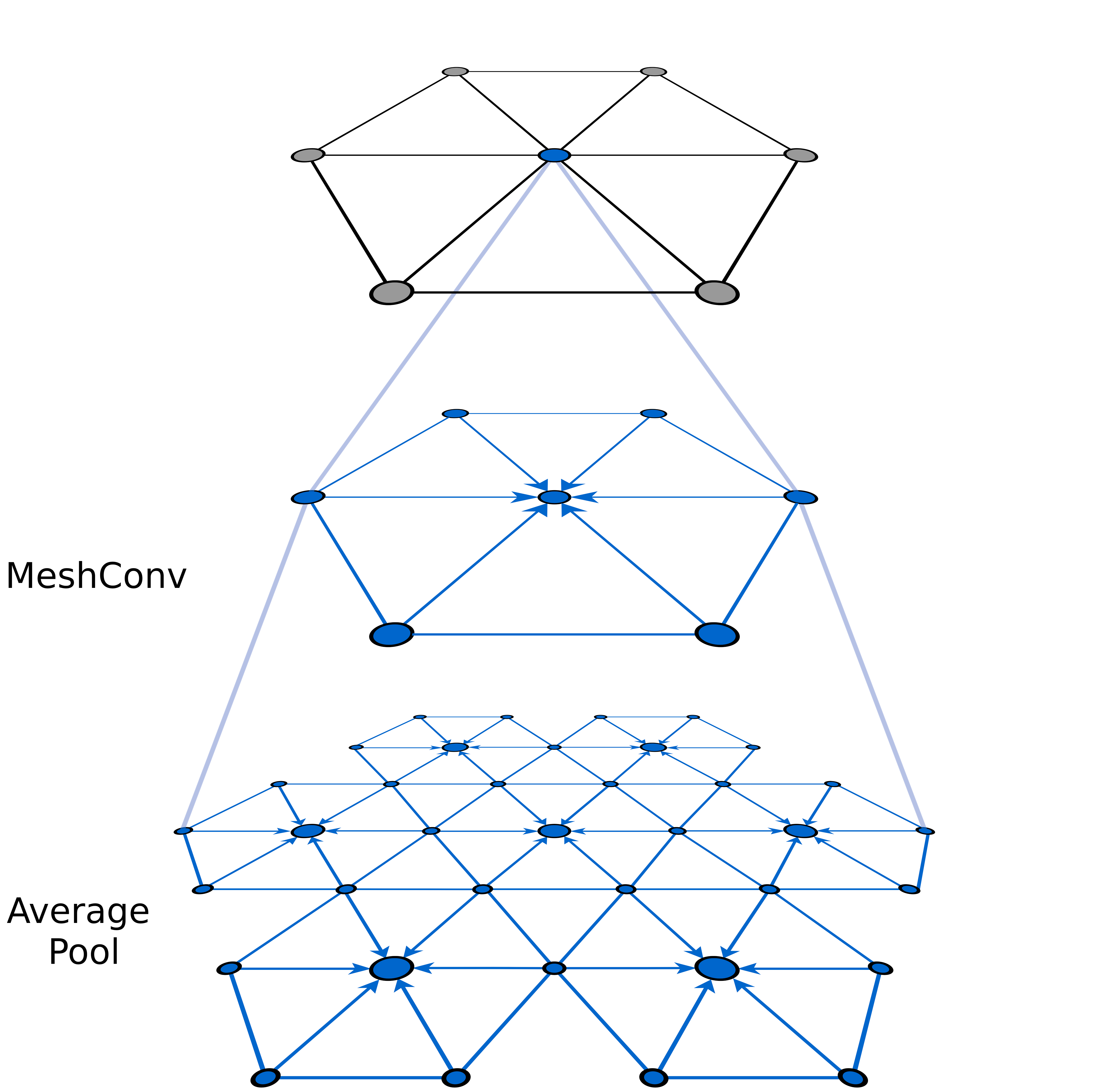}}}
    \caption{The receptive field across the downsampling block depends on the order of MeshConv and pooling operations. Swapping the order of operations changes the receptive field from a 3-ring to a 2-ring neighbourhood.}
    \label{fig:RF_hierarchy}
\end{figure}

These design choices have a direct effect on the receptive field of the MeshConv operation at each level, and therefore the receptive hierarchy our FPN is designed to exploit. Our choice of ordering and average pooling maintains the same receptive field as the seminal work, but without losing information in down-sampling. Alternatively, by keeping the original ordering with average pooling, the receptive field is larger by a one-ring neighborhood of the previous level, see figure \ref{fig:RF_hierarchy}. This will increase the receptive field at each level of the encoder, which may not be beneficial for finding smaller scale features. For this reason, we perform an ablation study over these design choices to empirically determine the optimal up/down-sampling procedure for spherical FPNs.
\section{Experiments}
\subsection{Stanford 2D-3D-S Experiments}
\label{subsec:2d3ds}
We follow \citet{jiang2019spherical} and analyse our models on the Stanford 2D3DS spherical image dataset, which contains 1413 equirectangular images of indoor scenes, with RGBD channels and semantic labels corresponding to 13 classes. We use the pre-processed data provided by \citet{jiang2019spherical} which samples the original images at the latitude-longitudes of the spherical mesh vertex positions spherical signal. The input RGB-D channels are interpolated using bilinear interpolation, and semantic labels are acquired using nearest-neighbor interpolation. Model performance is measured using two standard metrics - pixel-wise accuracy, and mean intersection-over-union (mIoU),

We test a range of models, varying the maximum depth of the feature pyramid. Besides simply providing models of varying complexity, we motivate this based on the extreme spatial coarseness of the lowest level meshes. Planar FPNs \cite{lin2017feature}, which operate on established image recognition benchmark datasets such as ImageNet \cite{deng2009imagenet} and MS COCO \cite{mscoco}, cap their lowest pyramid level to feature maps with a resolution of $20 \times 15$, or 300 pixels. In contrast, the icosahedral mesh at the lowest level-0 consists of a mere 12 vertices, roughly corresponding to a $3 \times 4$ 2D image. At this extreme of spatial coarseness, the potential gains from including increasingly lower pyramid levels may be non-trivial. Objects in the scene may not have the excessively large features which are best represented at this resolution. For this reason, we experiment with a range of pyramid depths. 

\textbf{Experimental Setup:} We test spherical FPNs with maximum level of 5, and minimum level in \{0,1,2,3\}. All models use the bilinear upsampling and average pooling blocks defined previously. For all experiments we use the Adam optimizer to train our networks for 100 epochs, with an initial learning rate of 0.01 and a step decay of 0.9 every 20 epochs. We use a batch size of 16 for all models except the L0:5 FPN, for which we use a batch size of 8 due to memory restrictions. All models were trained on Google Cloud using an NVIDIA Tesla T4. The results are shown in Table \ref{tab:fpns}
\begin{figure}
    \centering
    \includegraphics[width=\linewidth]{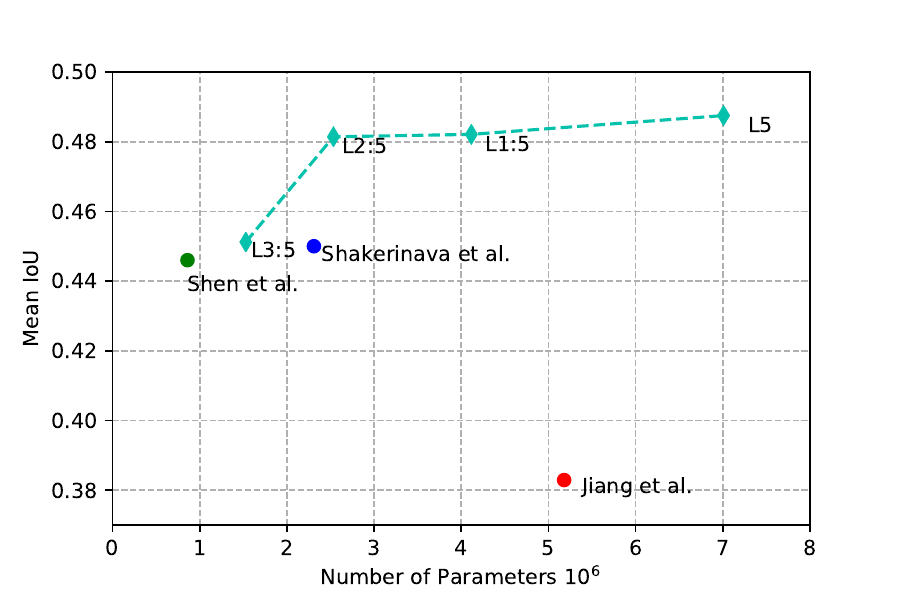}
    \caption{Mean IoU against the number of parameters for spherical FPNs (cyan diamond) of varying minimum levels.}
    \label{fig:iou_v_param}
\end{figure}

\begin{table}[ht]
\centering
\resizebox{\columnwidth}{!}{%
\begin{tabular}{cccc}
\hline
Model                 & Parameters        & Accuracy & mIoU   \\ \hhline{= = = =}
\cite{jiang2019spherical}                & 5.18M          & 54.65   & 0.3829 \\ 
\cite{shen2022pdoes2cnns}               & \textbf{0.86M}          & 60.4  & 0.4460 \\ 
\cite{pixelspheres}               & 2.31M         & 62.5  & 0.45 \\ \hline
\textbf{L3:5} $S^2$FPN                      & 1.53M    &  61.80  & 0.4512 \\ 
\textbf{L2:5} $S^2$FPN                      & 2.53M   & 63.83   & 0.4814 \\ 
\textbf{L1:5} $S^2$FPN                      & 4.12M   & 64.66  & 0.48.21 \\ 
\textbf{L0:5} $S^2$FPN                        & 7.01M    & \textbf{64.73}   & \textbf{0.4875} \\ \hline
\end{tabular}%
}
\caption{Results for FPN experiments with varied pyramid depth. L$x$:5 refers to a minimum level-$x$ mesh}
\label{tab:fpns}
\end{table}

\textbf{Results and Discussion:} All four FPN models achieve state-of-the-art performance, with the smallest L3:5 FPN using $1.5\times$ fewer parameters. The L2:5 model demonstrates substantial gains of 3.14 mIoU points over \cite{pixelspheres}, whilst using a comparable number of parameters to. Our deepest FPN achieves 48.75 mIOU, setting a new state-of-the-art for graph-based spherical CNNs on this dataset. The per-class mIoU scores are shown in Table \ref{tab:fpn_stanford_iou}. Collectively, Our $S^2$FPNs achieve the best performance across all the classes except door, where they fall short by 0.007 mIoU points.

\begin{table*}[ht]
\centering
\resizebox{\textwidth}{!}{%
\begin{tabular}{c|c|c|c|ccccccccccccc}
\hline
 & Model & Parameters & Mean & beam & board & bookcase & ceiling & chair & clutter & column & door & floor & sofa & table & wall & window \\ \hhline{= = = = = = = = = = = = = = = = = }
\multirow{2}{*}{Planar} & UNet & - &0.3587 & 0.0853 & 0.2721 & 0.3072 & 0.78757 & 0.3531 & 0.2883 & 0.0487 & 0.3377 & 0.8911 & 0.0817 & 0.3851 & 0.5878 & 0.2392 \\
 & FCN8s & - & 0.3560 & 0.0572 & 0.3139 & 0.2894 & 0.7981 & 0.3623 & 0.2973 & 0.0353 & 0.4081 & 0.8884 & 0.0263 & 0.3809 & 0.5849 & 0.1859 \\ \hline
\multirow{7}{*}{Spherical} &\cite{jiang2019spherical} & 5.18M & 0.3829 & 0.0869 & 0.3268 & 0.3344 & 0.8216 & 0.4197 & 0.2562 & 0.1012 & 0.4159 & 0.8702 & 0.0763 & 0.4170 & 0.6167 & 0.2349 \\\
 & \cite{shen2022pdoes2cnns} & \textbf{0.86M} & 0.446 & 0.114 & 0.433 & 0.382 & 0.839 & 0.503 & 0.313 & 0.124 & 0.484 & \textbf{0.90} & 0.181 & 0.495 & 0.659 & 0.371 \\
 & \cite{pixelspheres} &  2.31M & 0.450 & - & - & - & - & -& - & - & - & - & - & - & - & - \\\
 & \textbf{L3:5} $S^2$FPN & 1.53M & 0.451 & \textbf{0.140} & 0.391 & 0.395 & 0.848 & 0.536 & 0.326 & 0.142 & 0.470 & 0.890 & 0.171 & 0.512 & 0.658 & 0.387\\
 & \textbf{L2:5} $S^2$FPN & 2.53M & 0.481 & 0.126 & \textbf{0.509} & 0.423 & \textbf{0.850} & 0.563 & 0.350 & 0.149 & 0.500 & 0.892 & \textbf{0.240} & 0.543 & 0.686 & 0.426\\
 & \textbf{L1:5} $S^2$FPN & 4.12M & 0.482 & 0.128 & 0.499 & 0.436 & 0.849 & 0.546 & 0.352 & \textbf{0.171} & 0.511 & 0.893 & 0.238 & 0.538 & \textbf{0.692} & 0.415\\
 & \textbf{L0:5} $S^2$FPN & 7.01M & \textbf{0.487} & 0.09 & 0.505 & \textbf{0.443} & 0.842 & \textbf{0.581} & \textbf{0.352} & 0.158 & \textbf{0.535} & 0.893 & 0.237 & \textbf{0.558} & 0.691 & \textbf{0.451} \\ \hline
\end{tabular}%
}
\caption{Per-class mIoU on Stanford 2D-3D-S dataset}
\label{tab:fpn_stanford_iou}
\end{table*}

\subsection{Ablation Study}
We perform an ablation study over our choice of down-sampling and up-sampling operations, as well as the order of MeshConv operation relative to down-sampling. We consider two types of down-sampling to a lower-level mesh, average pooling vertex values to compute new vertices (``average"), or simply taking the values of vertices shared with the lower mesh (``drop"), as in \citet{jiang2019spherical}. For up-sampling, we test bilinear interpolation (``bilinear") and having new vertices take a value of zero (``zero-pad"). Finally, the condition ``swapped" refers to whether the order of MeshConv and downsampling operation is swapped relative to \citet{jiang2019spherical}. If the order is swapped, MeshConv is applied before downsampling, as in \ref{fig:RF_hierarchy}a). All models used a L3:5 architecture, and were tested using an identical experimental set up to the previous section.  
\begin{table}[ht]
\centering 
\resizebox{\columnwidth}{!}{%
\begin{tabular}{cccccc}
\hline
 Receptive Field & Swapped & Up-Sampling & Down-Sampling & Mean Acc (\%) & mIoU\\ \hhline{======}
2-ring & \xmark & zero-pad & drop & 59.5 & 0.422 \\ 
1-ring & \cmark & zero-pad & drop & 59.9 & 0.418 \\
1-ring & \cmark & bilinear & drop & 60.3 & 0.426 \\ 
2-ring & \cmark & zero-pad & average & 61.2 & 0.440 \\ 
3-ring & \xmark & bilinear & average & 60.5 & 0.442 \\
2-ring & \cmark & bilinear & average & \textbf{61.8} & \textbf{0.451} \\ \hline
\end{tabular}%
}
\caption{Ablation study over up/down-sampling design choices for a L3:5 FPN.}
\label{tab:fpns_ablation}
\end{table}

\textbf{Results and Discussion:} The use of bilinear up-sampling and average pooling is seen to improve the model both independently and in conjunction. Swapping the order of MeshConv and down-sampling operation is seen to harm performance for the model using "drop" down-sampling, but improve the model when applied to a model with average pooling. The reason for this requires more experimentation, but could be due to the fact that both swapping the order and using ``drop" down-sampling reduce the receptive field on the previous level mesh. More precisely, used together, they have a receptive field nine times as small as that of average pooling and the original ordering. Combined with using the shallowest L3:5 model, this significantly reduces receptive field on the input signal and may be harming the network performance.

\subsection{ClimateNet}
We also evaluate our method on the task proposed by \cite{Mudigonda2017SegmentingAT}, the segmentation of climate events from a 20-year run of the Community Atmospheric Model v5 (CAM5) \cite{climatedata}. We use the data preprocessed by \cite{jiang2019spherical}, which consists of spherical signals sampled onto a level-5 icosahedral grid. Each map consists of 16 channels of measurements such temperature, wind, humidity, and pressure. The training, validation, and test set size is 43917, 6275, and 12549, respectively. The task is to use these climate measurements to segment Atmospheric Rivers
(AR) and Tropical Cyclones (TC). The
labels are heavily unbalanced with 0.1\% TC, 2.2\% AR, and 97.7\% background (BG) pixels, and so we use a weighted cross-entropy loss. Please see \cite{Mudigonda2017SegmentingAT} for information on ground truth label production. Since the model proposed by \cite{jiang2019spherical} is our most direct comparison, we reduce the number of channels in our $S^2$FPN model (see \ref{fig:spherical_arch}) by a factor of 4, in order to have a comparable number of parameters. 

For all experiments we train for 50 epochs using the Adam optimizer, with an initial learning rate of 0.001, and a step decay of 0.4 every 20 epochs. We use a batch size of 128, training all models on a NVIDIA Tesla T4. 
\begin{table}[ht]
\centering
\resizebox{\columnwidth}{!}{%
\begin{tabular}{cccc}
\hline
Model                 & Parameters        & Mean Acc (\%) & mAP \\ \hhline{= = = =}
\cite{jiang2019spherical} & 330K         & 94.95   & 0.3841 \\
\cite{defferrard2020deepsphere} & 52M & 88.70 & 0.9138 \\
\cite{pixelspheres} & 2.31M & \textbf{99.30} & \textbf{0.9520} \\ \hline
\textbf{L1:5} $S^2$FPN   &  262K & 93.43 & 0.5448 \\ 
\textbf{L0:5} $S^2$FPN   &  444K & 94.10 & 0.5132 \\ \hline
\end{tabular}%
}
\caption{Results on ClimateNet}
\label{tab:fpns_climate}
\end{table}
We omit comparisons to \cite{defferrard2020deepsphere} and \cite{pixelspheres} since their parameter scales are vastly different ($52$M, and $2.3$M respectively). Compared to Spherical UNet, our FPN models show substantial improvements in terms of mAP. The L1:5 model performs better than the larger L0:5 model, perhaps indicating the effect of diminishing returns on extremely coarse meshes. 

\section{Conclusion}
At present, there is not a universally accepted approach for performing convolution on a sphere. Generalising convolutions to work on arbitrary shapes and unstructured grids is an active research topic. However, the importance of multi-scale features is generic. In theory, feature pyramids can be constructed using any network with semantic and receptive field hierarchies. In this work, we generalise the idea of feature pyramids to the spherical domain and propose an alternative model to UNet for spherical image segmentation. In addition, motivated by their influence on the receptive field hierarchy of the network, we present improved pooling and up-sampling schemes and measure their respective contributions through an ablation study. Our work demonstrates that FPNs are a powerful domain-agnostic architecture, which successfully generalize to a PDO-based spherical CNNs, attaining state-of-the-art results. 
\section{Acknowledgments}
We would like to thank Chiyu "Max" Jiang for his valuable comments on the up/down-sampling methods, and for their work on pre-processing both ClimateNet and Stanford 2D-3D-S datasets. 
\label{sec:reference_examples}

\bibliography{aaai22}

\end{document}